\documentclass{ceurart}       
\usepackage[utf8]{inputenc}
\usepackage[english]{babel}
\usepackage{microtype}
\usepackage{csquotes}
\usepackage{textcomp}        
\usepackage{newunicodechar}  

\newunicodechar{§}{\S}     
\newunicodechar{÷}{\textdiv} 
\newunicodechar{×}{\texttimes}
\newunicodechar{±}{$\pm$}   

\usepackage{amsmath}
\usepackage{graphicx}
\usepackage{booktabs}
\usepackage{tabularx}
\usepackage{subcaption}
\usepackage{pgfplots}
\pgfplotsset{compat=1.18}

\usepackage{algorithm}
\usepackage[noend]{algpseudocode}
\usepackage{listings}
\usepackage{xcolor}

\usepackage{tikz}
\usepackage{enumitem}
\usepackage{placeins}
\usepackage{tcolorbox}

\definecolor{codegreen}{rgb}{0,0.6,0}
\definecolor{codegray}{rgb}{0.5,0.5,0.5}
\definecolor{codepurple}{rgb}{0.58,0,0.82}
\definecolor{backcolour}{rgb}{0.97,0.97,0.95}
\definecolor{promptblue}{rgb}{0.1,0.1,0.7}
\definecolor{promptorange}{rgb}{0.8,0.4,0}
\definecolor{rulecolor}{rgb}{0.85,0.85,0.85}
\definecolor{keywordcolor}{rgb}{0.1,0.1,0.7}
\definecolor{commentcolor}{rgb}{0.3,0.5,0.3}
\definecolor{stringcolor}{rgb}{0.6,0.2,0.2}

\lstdefinestyle{mystyle}{
  backgroundcolor=\color{backcolour},
  basicstyle=\footnotesize\ttfamily,
  breaklines=true,
  frame=single,
  framerule=0.4pt,
  keywordstyle=\color{keywordcolor}\bfseries,
  commentstyle=\color{codegreen}\itshape,
  stringstyle=\color{codepurple},
  numberstyle=\tiny\color{codegray},
  showstringspaces=false,
  tabsize=2
}

\lstdefinestyle{MsftResearchStyle}{
  basicstyle=\footnotesize\ttfamily,
  breaklines=true,
  frame=tb,
  framerule=0.4pt,
  rulecolor=\color{rulecolor},
  keywordstyle=\color{keywordcolor}\bfseries,
  commentstyle=\color{commentcolor}\itshape,
  stringstyle=\color{stringcolor},
  showstringspaces=false,
  columns=fullflexible
}

\newcommand{\thinktag}{\textcolor{promptblue}{\texttt{<think>}}\xspace}
\newcommand{\closethinktag}{\textcolor{promptblue}{\texttt{</think>}}\xspace}
\newcommand{\querytag}{\textcolor{promptorange}{\texttt{<query>}}\xspace}
\newcommand{\closequerytag}{\textcolor{promptorange}{\texttt{</query>}}\xspace}
\newcommand{\queryresulttag}{\textcolor{codegreen}{\texttt{<query\_result>}}\xspace}
\newcommand{\closequeryresulttag}{\textcolor{codegreen}{\texttt{</query\_result>}}\xspace}
\newcommand{\answertag}{\textcolor{codepurple}{\texttt{<answer>}}\xspace}
\newcommand{\closeanswertag}{\textcolor{codepurple}{\texttt{</answer>}}\xspace}

\newtcolorbox{turnbox}[1][]{
  enhanced, breakable,
  colback=white, colframe=black!15,
  fonttitle=\small\bfseries\sffamily,
  colbacktitle=black!5,
  title=#1,
  boxsep=3pt, arc=2mm, boxrule=0.6pt
}
\newcommand{\turnseparator}{\par\color{black!20}\rule{\linewidth}{0.2pt}\par}

\begin{document}

\copyrightyear{2025}
\copyrightclause{Copyright for this paper by its authors. Use permitted under Creative Commons License Attribution 4.0 International (CC BY 4.0).}

\conference{RAGE-KG 2025: The Second International Workshop on Retrieval-Augmented Generation Enabled by Knowledge Graphs, co-located with ISWC 2025, November 2--6, 2025, Nara, Japan}

\title{Learning to Refine: An Agentic RL Approach for Iterative SPARQL Query Construction}

\author[1,2]{Floris Vossebeld}[
  orcid=0000-000X-XXXX-XXXX,
  email=f.r.vossebeld@student.utwente.nl,
]
\author[1]{Shenghui Wang}[
  orcid=0000-0003-0583-6969,
  email=shenghui.wang@utwente.nl
]

\address[1]{Faculty of Electrical Engineering, Mathematics and Computer Science, 
            University of Twente, Drienerlolaan 5, 7522 NB Enschede, The Netherlands}

\address[2]{Microsoft Netherlands,
            Evert van de Beekstraat 354, 1118 CZ Schiphol, The Netherlands}

\fntext[1]{Work performed while the author was an Intern at
Microsoft Netherlands (Feb-Jul 2025).  
The views expressed are the author’s and do not necessarily reflect those of Microsoft Corporation.}

\renewcommand{\shortauthors}{F. Vossebeld}

\begin{abstract}
Generating complex, logically-sound SPARQL queries for multi-hop questions remains a critical bottleneck for Knowledge Graph Question Answering, as the brittle nature of one-shot generation by Large Language Models (LLMs) hinders reliable interaction with structured data. Current methods lack the adaptive policies needed to dynamically debug queries based on real-time execution feedback. This paper introduces a novel agentic framework where an LLM learns a resilient policy for the sequential process of iterative SPARQL construction. We show that a compact 3B-parameter model, trained exclusively via outcome-driven Reinforcement Learning (GRPO) without supervised fine-tuning, can learn effective policies for this task, discovering how to systematically recover from execution errors and refine its queries toward a correct answer. On a curated, executable single-answer subset of LC-QuAD 2.0, our agent achieves 49.7\% accuracy post-entity-linking, a significant 17.5 percentage point improvement over the strongest iterative zero-shot baseline. Further analysis reveals that while the agent's capability is driven by RL, its performance is enhanced by an explicit deliberative reasoning step that acts as a cognitive scaffold to improve policy precision. This work presents a generalizable blueprint for teaching agents to master formal, symbolic tools through interaction, bridging the gap between probabilistic LLMs and the structured world of Knowledge Graphs.

\end{abstract}


\maketitle
\section{Introduction}
\label{sec:introduction}
Knowledge Graphs (KGs) like DBpedia \citep{lehmann_dbpedia_2015} and Wikidata \citep{vrandecic_wikidata_2014} structure vast information as entities linked by relations \citep{hogan_knowledge_2022}. Answering natural language questions using these graphs, Knowledge Graph Question Answering (KGQA), is vital for applications from search to decision support \citep{noy_industry-scale_2019, zhang_variational_2017}. While single-hop KGQA is well studied, multi-hop questions, requiring reasoning across multiple entities and relations, remain a major challenge due to combinatorial path explosion \citep{sun_pullnet_2019}, KG incompleteness \citep{ren_lego_2021}, and semantic ambiguity \citep{lin_kagnet_2019}.

Traditional KGQA methods such as semantic parsing \citep{berant_semantic_2013}, retrieval-based approaches \citep{sun_open_2018}, and KG embeddings \citep{bordes_translating_2013} often apply static reasoning strategies. These methods typically generate a SPARQL query in a single pass or retrieve a fixed subgraph, lacking mechanisms for iterative refinement. As a result, they struggle with long or error-prone queries, and fail to recover from intermediate execution failures.

Large Language Models (LLMs) have recently shown strong capabilities in structured reasoning, particularly when supported by techniques like Chain-of-Thought prompting \citep{wei_chain--thought_2023}. Agentic frameworks such as ReAct \citep{yao_react_2023} and StructGPT \citep{jiang_structgpt_2023} allow LLMs to combine reasoning with tool use. However, many of these methods either rely on predefined tools or prompting schemes, and do not learn adaptive interaction policies through feedback.

Consider the question \enquote{Which actors starred in movies directed by the director of \emph{Inception}?} Answering this requires identifying the film, finding its director, retrieving the director’s other films, and then the actors in those films, each step relying on correct schema navigation, relation selection, and entity disambiguation. Translating this full path into a correct SPARQL query in one shot is error-prone. An alternative is to incrementally build and test parts of the query, adapt based on results, and correct mistakes mid-way, requiring interaction with the KG as a semantic environment.

While integrating LLMs with KGs is actively researched \citep{pan_unifying_2024, chakraborty_multi-hop_2024, jiang_structgpt_2023}, methods often rely on fixed interaction logic, prompting strategies applied to base models, or using KG information primarily as retrieved context. They typically do not involve fine-tuning the model specifically to learn adaptive policies for the iterative construction of structured SPARQL queries based on execution feedback, which is crucial for handling the complexity and potential errors inherent in multi-step KG interactions.

This paper addresses this gap by proposing an agentic framework in which a language model learns a policy for iterative SPARQL query construction through interaction with a knowledge graph. The agent operates in a think–act–observe loop: it reasons about the current state (\thinktag), generates a SPARQL query or final answer (\querytag, \answertag), and receives execution feedback from the KG (\queryresulttag). Rather than relying on static, one-shot generation, the agent adapts its strategy based on results, including errors or empty outputs, progressively refining its queries. To enable this behavior, we fine-tune a compact LLM using Group Relative Policy Optimization (GRPO), a reinforcement learning algorithm designed for sparse, outcome-based rewards. The agent learns not only to generate queries, but to interpret feedback and dynamically debug or explore, improving robustness in complex multi-hop scenarios.

This motivates the following research questions: 
\begin{tcolorbox}[
  title=Research questions,
  colback=white,
  colframe=black!30,
  boxrule=0.45pt,
  arc=1.2mm,
  left=2.5mm,
  right=2.5mm,
  top=1mm,
  bottom=1mm]
\begin{enumerate}[leftmargin=2.5em,label=\textbf{RQ\arabic*:},itemsep=3pt]
  \item How can an LLM learn to \emph{iteratively} build and refine SPARQL
        queries using execution feedback to answer complex multi-hop KG
        questions?
  \item Can reinforcement learning effectively train such an agent to produce
        accurate answers from outcome signals alone?
  \item How does this iterative, RL-guided approach compare with static or
        prompt-only baselines on the LC-QuAD 2.0 benchmark?
\end{enumerate}
\end{tcolorbox}



The remainder of this thesis details related work (\S\ref{sec:related-work}), presents our methodology (\S\ref{sec:methodology}), details the experiments and results (\S\ref{sec:experiments} and \S\ref{sec:results}), and discusses the findings and discussion and conclusions (\S\ref{sec:discussion-conclusion}).

\section{Related work}
\label{sec:related-work}

Multi-hop KGQA requires combining structured reasoning, language understanding, and interaction. We review prior work on traditional KGQA methods, agentic LLMs, and reinforcement learning, highlighting how our approach integrates symbolic interaction, tool-based reasoning, and adaptive query construction to address the unique challenges of multi-hop KGQA.

\paragraph{Multi-hop KGQA and traditional approaches}

Knowledge Graph Question Answering (KGQA) maps natural language questions to structured answers by reasoning over triples in a knowledge graph (KG) \citep{zhang_variational_2017, sun_open_2018}. While single-hop questions can be resolved through direct relations, multi-hop KGQA requires compositional reasoning across multiple entities and relations \citep{saxena_improving_2020, gu_beyond_2021}. This increases complexity due to path explosion \citep{sun_pullnet_2019}, KG incompleteness and noise \citep{ren_lego_2021}, and semantic ambiguity \citep{lin_kagnet_2019}.

Traditional KGQA methods fall into three categories: semantic parsing, retrieval-based approaches, and embedding-based reasoning. Semantic parsing methods aim to generate formal queries such as SPARQL \citep{berant_semantic_2013, yih_semantic_2015}, but are brittle to linguistic variation and require substantial supervision \citep{liang_learning_2011, yih_value_2016}. Retrieval-based methods extract subgraphs for ranking \citep{sun_open_2018, sun_pullnet_2019} but often struggle with complex logic. Embedding-based approaches reason in vector space \citep{bordes_translating_2013, saxena_improving_2020}, sacrificing interpretability and logical precision. Critically, these approaches apply fixed computation and lack iterative refinement mechanisms based on intermediate feedback, hindering performance on complex multi-hop tasks.

\paragraph{Agentic LLMs and symbolic interaction in KGQA}
\label{subsec:agentic-llms}

Recent work leverages LLMs as agents that combine internal reasoning with external actions. Agentic frameworks such as ReAct \citep{yao_react_2023} and MRKL \citep{karpas_mrkl_2022} allow LLMs to operate in a loop of \texttt{<think> → <act> → <observe>}, interacting with tools to solve complex tasks. In KGQA, systems like StructGPT \citep{jiang_structgpt_2023} and Think-and-Graph \citep{sun_think--graph_2024} extend this idea by giving LLMs access to navigation tools (e.g., retrieving neighbors or relations). However, these tools are often predefined, and reasoning policies are static or heuristic-driven, limiting adaptivity.

Our work shifts the focus from tool-based navigation to formal query generation. Inspired by ARTIST \citep{singh_agentic_2025}, we treat SPARQL construction as the agent’s primary action. The model alternates between \texttt{\thinktag}, \texttt{\querytag}, and \texttt{\answertag} tags, learning to refine its reasoning through symbolic interaction with the KG. This reframes KGQA as a dynamic decision-making process grounded in executable feedback.

This strategy also relates to test-time compute scaling \citep{snell_scaling_2024}, where additional reasoning effort is allocated adaptively. Some approaches use inference-time sampling or search \citep{wang_self-consistency_2023, yao_tree_2023}; others explicitly train models to optimize reasoning under compute constraints \citep{shao_deepseekmath_2024, lightman_lets_2023}. Our work falls in the latter category, focusing on training an agent to effectively use interaction cycles for symbolic query refinement.

\paragraph{Reinforcement learning for iterative query generation}

While supervised fine-tuning enables LLMs to imitate reasoning (e.g., Chain-of-Thought prompting \citep{chung_scaling_2022}), it relies heavily on high-quality demonstrations and struggles with long-horizon credit assignment. Reinforcement learning (RL) offers a more flexible alternative, enabling agents to learn from outcome-based interaction.

We build on \textit{Group Relative Policy Optimization (GRPO)}, a recent RL algorithm designed for sparse, symbolic environments. GRPO has shown success in math problem solving \citep{shao_deepseekmath_2024}, SQL generation \citep{ma_sql-r1_2025}, and general tool use \citep{singh_agentic_2025}. In our setting, GRPO allows a compact LLM to learn symbolic refinement strategies from task-level rewards alone, recovering from syntax errors, adapting query structure, and issuing exploratory probes. This enables robust multi-hop reasoning without requiring step-level supervision or hand-coded recovery heuristics.

\section{Methodology}
\label{sec:methodology}

Our approach transforms multi-hop KGQA from a one-shot generation task into an iterative, sequential decision-making problem. We developed an autonomous agent, powered by a Large Language Model (LLM), that learns an optimal policy for constructing and refining SPARQL queries through live interaction with a Knowledge Graph (KG). The agent operates within a Reinforcement Learning (RL) framework, where its behavior is optimized to maximize a reward signal reflecting the accuracy and validity of its actions.

\begin{figure}[t]
  \centering
  \includegraphics[width=\textwidth]{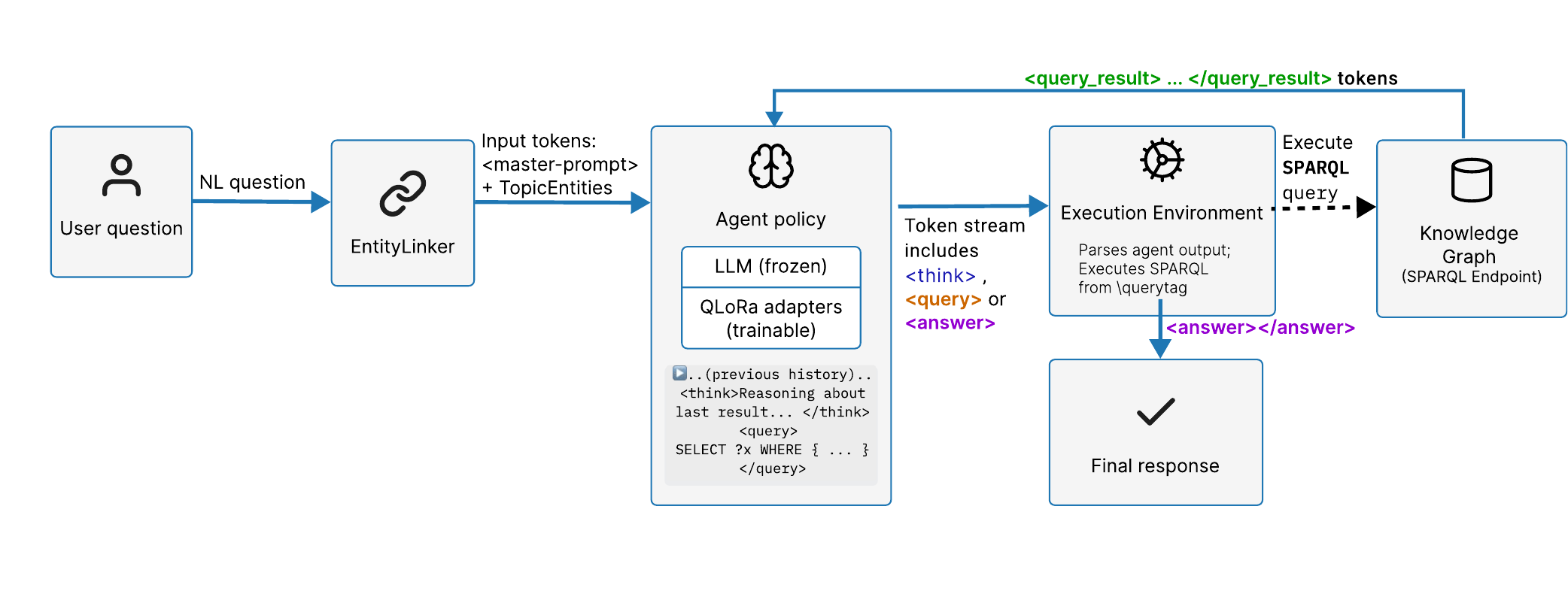}
  \caption{The end-to-end agentic inference loop. The agent policy (LLM + QLoRA adapters) receives the current state (question and history) and generates reasoning (\thinktag) followed by an action: either a SPARQL query (\querytag) or the final answer (\answertag). The SPARQL query is executed against the KG, with the outcome (\queryresulttag) updating the state for the next iteration. The loop terminates when the agent produces an answer.}
  \label{fig:inference_loop}
\end{figure}

The agent's core is an interaction loop, conceptually illustrated in Figure \ref{fig:inference_loop}. In each turn, the agent: 1) \textbf{analyzes} the history of the task, including the initial question and all previous KG interactions; 2) \textbf{reasons} about the next best step within a \thinktag block; and 3) \textbf{acts} by generating either a new SPARQL query (\querytag) or a final answer (\answertag). This cycle repeats until the agent confidently terminates the process. This section details the formal problem definition, the mechanics of the agent-environment interaction, and the RL-based training process used to learn the query refinement policy.
\begin{table}[t]
    \centering
    \caption{Key components of the agentic KGQA framework.}
    \label{tab:methodology_components}
    \begin{tabularx}{\columnwidth}{@{} l X @{}}
        \toprule
        \textbf{Component} & \textbf{Role and implementation} \\
        \midrule
        \textbf{Agent policy} ($\pi_{\theta}$) & An instruction-tuned LLM (\texttt{Qwen2.5-3B-Instruct}) with QLoRA adapters ($\theta$) that generates think-act sequences. \\
        \addlinespace
        \textbf{Environment} & Executes SPARQL queries against a self-hosted Wikidata endpoint and returns structured feedback (\texttt{\queryresulttag}). \\
        \addlinespace
        \textbf{State} ($s_t$) & The full, structured conversation history, including all prior agent actions and environment observations. \\
        \addlinespace
        \textbf{Action} ($a_t$) & A text generation containing a \thinktag block followed by either a \querytag or \answertag block. \\
        \addlinespace
        \textbf{Reward} ($R(\tau)$) & A terminal, composite reward signal evaluating structural validity, final answer correctness, and efficiency. \\
        \addlinespace
        \textbf{Learning algorithm} & Group Relative Policy Optimization (GRPO) to fine-tune the policy $\pi_{\theta}$ based on the outcome-based reward $R(\tau)$. \\
        \bottomrule
    \end{tabularx}
\end{table}

\subsection{Formalism: an agentic Markov Decision Process}
\label{subsec:methodology_mdp}
We model the iterative query construction task as a finite-horizon Markov Decision Process (MDP), defined by the tuple $(\mathcal{S}, \mathcal{A}, \mathcal{P}, \mathcal{R}, \gamma)$.
\begin{description}
    \item[State ($s_t \in \mathcal{S}$)] A state is the full conversation history at turn~$t$. It is a structured text sequence comprising the initial prompt, the user's question, and all subsequent agent turns and environment observations: $s_t = (\text{prompt}, m_1, o_1, m_2, o_2, \ldots)$.
    
    \item[Action ($a_t \in \mathcal{A}$)] An action is the complete text sequence generated by the agent in a single turn. It must conform to one of two valid structures: a reasoning block followed by a query, or a reasoning block followed by a final answer.
    \[ a_t = 
        \begin{cases}
          \texttt{\thinktag...\closethinktag\querytag...\closequerytag} \\
          \texttt{\thinktag...\closethinktag\answertag...\closeanswertag}
        \end{cases}
    \]
    The action space $\mathcal{A}$ is the vast set of all possible text generations that adhere to this format.
    
    \item[Transition function ($\mathcal{P}$)] The state transition $P(s_{t+1} \mid s_t, a_t)$ is largely deterministic. Given state $s_t$ and a query action $a_t$, the environment executes the SPARQL query from $a_t$. The resulting observation $o_{t+1}$ (the content of the \queryresulttag block) is appended to the history to form the next state $s_{t+1} = s_t \circ a_t \circ o_{t+1}$. The episode terminates if the agent produces an answer action, exceeds the maximum number of turns, or generates a malformed action.
    
    \item[Reward function ($\mathcal{R}$)] The reward $R(\tau)$ is a terminal, outcome-based reward assigned at the end of a full trajectory $\tau$. It is a composite signal designed to evaluate the success of the agent’s multi-turn strategy, as detailed in Section~\ref{subsec:grpo_reward_design}.
    
    \item[Policy ($\pi_{\theta}$)] The agent’s policy is the LLM itself, parameterized by a set of trainable QLoRA adapter weights $\theta$. The policy $\pi_{\theta}(a_t \mid s_t)$ maps the current state (history) to a probability distribution over the action space. Our objective is to find the optimal weights $\theta^*$ that maximize the expected terminal reward.
\end{description}
\subsection{Agent–environment loop}
\label{subsec:methodology_interaction_loop}

\begin{tcolorbox}[
  title=One interaction turn,
  colback=white,
  colframe=black!30,
  boxrule=0.45pt,
  arc=1.2mm,
  left=2.5mm,
  right=2.5mm,
  top=1mm,
  bottom=1mm]
\begin{enumerate}[leftmargin=1.3em,itemsep=2pt]
  \item \textbf{Think $\rightarrow$ Act}\,:  
        $\pi_\theta$ appends a \texttt{\thinktag} block \emph{plus} either  
        \texttt{\querytag} (SPARQL) or \texttt{\answertag}.
  \item \textbf{Environment}\,:  
        \begin{enumerate}[nosep,label*=\arabic*.]
          \item \texttt{\answertag} $\Rightarrow$ episode ends.  
          \item \texttt{\querytag} $\Rightarrow$ KG executes; reply comes back as \texttt{\queryresulttag}.  
          \item malformed output $\Rightarrow$ abort and error flag.
        \end{enumerate}
  \item Loop until success or turn limit $T_{\max}$.
\end{enumerate}
\end{tcolorbox}

\subsubsection{Knowledge‐graph execution environment}
\label{subsubsec:kg_environment_final}
For RL we require a fast, quota-free SPARQL endpoint. We therefore deploy a
containerised \emph{qEndpoint} (truthy Wikidata HDT) inside our Azure VNet.  A lightweight
\texttt{aiohttp} client issues queries asynchronously, with an in-memory LRU
cache, pre-flight syntax checks (\texttt{rdflib}), and automatic retry/back-off.
The result is a private, low-latency endpoint that sustains the thousands of
queries demanded by training.

\subsubsection{Agent Prompting and Structured Actions.}
The agent's policy is guided by a detailed system prompt, which provides task instructions, defines the required interaction format, and includes few-shot examples of successful refinement trajectories. 

A critical technique during RL training is \textbf{loss masking}. The policy's parameters $\theta$ are updated only based on the log-probabilities of tokens generated by the agent (i.e., within \thinktag, \querytag, and \answertag). Tokens from the environment (the initial prompt and all \queryresulttag blocks) are masked out from the loss calculation. This follows best practices from frameworks like ARTIST \citep{singh_agentic_2025} and focuses the learning signal squarely on the agent's decision-making policy, rather than wasting capacity trying to predict deterministic environment outputs.

\subsection{Policy optimization via Group Relative Policy Optimization (GRPO)}
\label{subsec:grpo_finetuning_revised}
We used Reinforcement Learning to fine-tune the agent's policy, specifically choosing Group Relative Policy Optimization (GRPO) \citep{shao_deepseekmath_2024}. This allows the agent to learn complex, sequential strategies from interactive experience and sparse, outcome-based rewards.

We fine-tuned the agent’s policy using Group Relative Policy Optimization (GRPO) \citep{shao_deepseekmath_2024}, a reinforcement learning algorithm well-suited to sparse, outcome-based tasks. GRPO compares the terminal reward of each trajectory against others in a group sampled from the same prompt, using relative performance to compute an advantage signal. This enables learning effective query refinement strategies without requiring a learned value function.

\subsubsection{Reward design for effective learning}
\label{subsec:grpo_reward_design}

\begin{tcolorbox}[
  title=Terminal reward $R(\tau)$,
  colback=white,
  colframe=black!30,
  boxrule=0.45pt,
  arc=1.2mm,
  left=2.5mm,right=2.5mm,top=1mm,bottom=1mm]

\[
R(\tau)=
\begin{cases}
-1, & \text{if trajectory \emph{not} structurally valid},\\[6pt]
1 \;+\; R_{\text{ans}}(\tau)\;-\;\bigl(0.1\,N_{\text{err}} + 0.02\,T\bigr),
  & \text{otherwise.}
\end{cases}
\]

\medskip
\begin{tabularx}{\linewidth}{@{}l X@{}}
\textbf{Structural validity} &
correct tag format and termination with \texttt{\answertag}. \\[4pt]
\textbf{Answer term} &
$R_{\text{ans}}(\tau)=
\begin{cases}
+0.5 & \text{if judge deems the answer correct},\\
-0.2 & \text{otherwise}
\end{cases}$ \\[8pt]
\textbf{Cost term} &
$0.1\,N_{\text{err}} + 0.02\,T$, where $N_{\text{err}}$ is the number of
failed SPARQL executions and $T$ the number of agent turns. \\
\end{tabularx}
\end{tcolorbox}

\subsubsection{Training Protocol and Implementation.}
\label{subsec:training_protocol}
\begin{figure}[t]
  \centering
  \includegraphics[width=\textwidth]{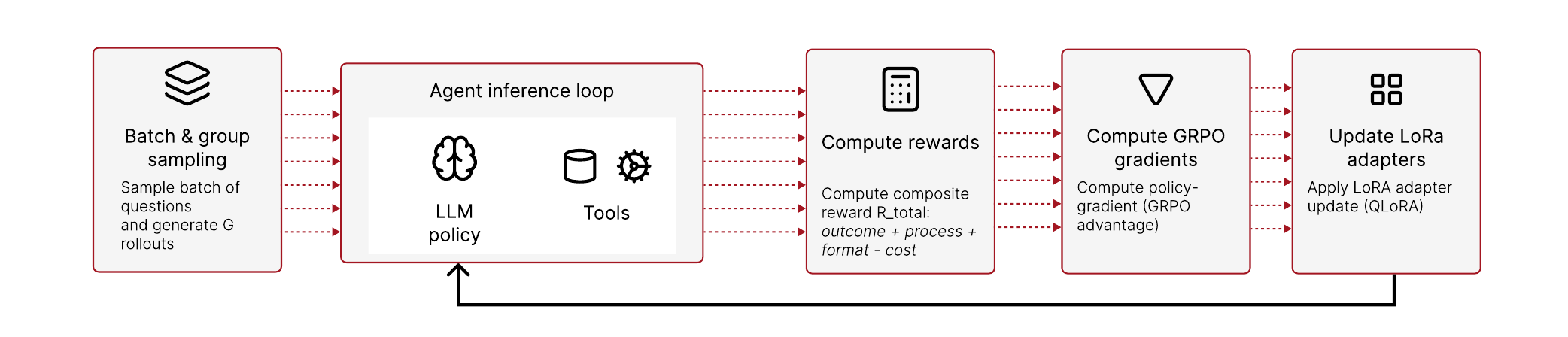}
  \caption{The end-to-end RL fine-tuning cycle. A batch of questions is sampled, and for each, the agent (LLM + LoRA adapters) generates $G$ rollouts using the iterative inference loop. The composite reward $R(\tau)$ is computed for each trajectory. GRPO uses these rewards to calculate a policy gradient, which is then used to update the LoRA adapter weights.}
  \label{fig:training_loop}
\end{figure}

The policy $\pi_{\theta}$ was fine-tuned using QLoRA \citep{dettmers_qlora_2023} with the Unsloth library's optimizations for memory and speed. The training, depicted in Figure \ref{fig:training_loop}, was executed on an Azure ML compute cluster with NVIDIA H100 GPUs. The GRPO training loop proceeds as follows:
\begin{enumerate}
    \item \textbf{Rollout generation:} Sample a batch of questions. For each question, generate $G=16$ full agentic trajectories using the current policy $\pi_{\theta}$.
    \item \textbf{Reward calculation:} Compute the composite reward $R(\tau)$ for each of the $G$ trajectories.
    \item \textbf{Policy update:} Use the GRPO objective to calculate the policy gradient, where trajectories with a reward greater than their group's average contribute positively.
    \item \textbf{Parameter update:} Update the LoRA adapter weights $\theta$ via the AdamW optimizer, applying loss masking and KL-divergence regularization to maintain stability.
\end{enumerate}
This cycle repeats, progressively improving the agent's policy. 

\subsection{Design choices and scope}
\label{subsec:methodology_choices}
Several key design choices shaped this research. A key design choice was to proceed directly to RL fine-tuning, deliberately bypassing a supervised fine-tuning (SFT) phase on gold trajectories. This tests a critical hypothesis for the field: \textit{can the combination of a powerful base model's instruction-following ability and a well-designed RL reward signal be sufficient to learn complex, tool-using behaviors, thereby reducing the dependency on costly, expert-curated demonstration data?} Second, our reward function's heavy penalty for structural and execution errors was a deliberate choice to force the agent to prioritize generating valid and executable SPARQL above all else. Finally, we must acknowledge two critical scope limitations that bound our claims: our system's performance is evaluated on a curated subset of single-answer questions, and it relies on pre-linked entities provided in the dataset. We did not address the significant challenges of entity linking or multi-answer aggregation, which remain out of scope for this thesis.

\section{Experimental Setup}
\label{sec:experiments}
This section outlines the experimental setup used to evaluate our agentic reinforcement learning approach to multi-hop KGQA. We describe the dataset curation process, model configuration, and knowledge graph environment, followed by training details and a report on compute and energy usage for reproducibility. We then present the comparative baselines and the evaluation methodology used to assess performance.


\subsection{Dataset}
\label{subsec:dataset}

Following \S\ref{subsubsec:kg_environment_final}, we re-execute every
gold query of LC-QuAD 2.0 against the frozen 2023-12-21 Wikidata HDT
dump and keep a triple $\langle q,\,query,\,answer\rangle$ only if the
query (i) succeeds, (ii) returns exactly one row, and (iii) yields a
valid RDF term.  The resulting corpus preserves entity, literal, and
boolean answers while discarding noisy items (Table~\ref{tab:dataset_stats}).

\begin{table}[h]
  \centering
  \caption{Curated single-binding subset of LC-QuAD 2.0.}
  \label{tab:dataset_stats}
  \begin{tabular}{lrr}
    \toprule
    \textbf{Split} & \textbf{Original size} & \textbf{Curated size} \\
    \midrule
    Train & 19\,344 & 5\,112 \\
    Test  &  4\,836 & 1\,279 \\
    \bottomrule
  \end{tabular}
\end{table}

\subsection{Core models and KG environment}
\label{subsec:model_environment}

We use the \texttt{unsloth/Qwen2.5-3B-Instruct-bnb-4bit} model\footnote{Model commit ID \texttt{2672b58} on the HuggingFace Hub}, selected for its instruction-following quality and compatibility with 4-bit QLoRA fine-tuning. Parameter-efficient training is performed using Unsloth’s QLoRA implementation with commonly used hyperparameters: rank 64, $\alpha=16$, dropout 0.05, learning rate $5\times10^{-6}$, group size $G=16$, KL coefficient $\beta=0.04$, and batch size 128. These values were selected through light, manual trial-and-error
only; No systematic hyperparameter tuning was conducted.

The agent interacts with a private SPARQL endpoint (qEndpoint v2.5.2) loaded with the 2023-12-21 “truthy” HDT dump of Wikidata. All SPARQL queries are executed asynchronously with memoization, exponential backoff, and a 3-second timeout.


\subsection{Training and compute setup}
\label{subsec:training_compute}

We fine-tune the agent using Group Relative Policy Optimization (GRPO), a reinforcement learning algorithm well-suited for sparse, outcome-based rewards. At each update step, $G = 16$ full roll-outs are sampled for each of 128 training questions. Terminal rewards are computed based on final answer correctness (see Section~\ref{subsec:grpo_reward_design}), and the LoRA weights are optimized using AdamW with a learning rate of $5 \times 10^{-6}$ and KL coefficient $\beta = 0.04$. Each interaction episode is capped at ten \texttt{<think>}–\texttt{<query>} cycles, and individual SPARQL executions are limited to a 3-second timeout.

Training is performed over a single epoch, converging in 11.5 hours on an NVIDIA H100 GPU (94 GB). This process consumed approximately 4.6 kWh of energy, which corresponds to an estimated 1.7 kg CO\textsubscript{2}e under the 2024 Dutch grid emission factor (0.37 kg CO\textsubscript{2}e/kWh). While the model is relatively compact, reinforcement learning remains computationally intensive, and further work is needed to evaluate the scalability and energy efficiency of this approach at larger scales or across multiple domains.


\subsection{Comparative baselines}
\label{subsec:comparative_baselines}
We compare our agentic RL model to three baselines:

\begin{description}
  \item[B1: Direct QA (Zero-Shot CoT)]  
        The base model answers from its parametric knowledge only, prompted with two chain-of-thought exemplars.
  \item[B2: One-Shot SPARQL]  
        A single-turn prompt instructs the model to emit a full SPARQL query; decoding uses temperature 0.2 and top-p 0.95.
  \item[B3: Prompt-Guided Iterative Agent]  
        Our think-query loop without RL; identical prompt as the RL-tuned agent and greedy decoding.
\end{description}


\subsection{Evaluation protocol and metrics}
\label{subsec:evaluation_protocol}


We evaluate KG-based agents using three key end-to-end metrics: answer accuracy, query executability, and interaction length. 

\begin{description}
\item[Accuracy]  
Correctness is determined by a frozen LLM-based evaluator (\texttt{GPT-4o-nano}) shared across all systems, including Direct QA. The evaluator receives the question, gold scalar binding, and the model’s \texttt{\answertag} response, and returns a Boolean verdict along with a justification. This allows for semantically equivalent but non-identical answers—e.g., paraphrasing, formatting differences, or unit conversions—to be marked as correct, unlike exact string matching.

\item[Executability rate] 
The proportion of all SPARQL queries generated by a system that are syntactically valid and execute successfully against the KG—computed as total successful executions divided by the total number of queries issued across the test set.

\item[Average turns]  
The mean number of agent interaction steps per question. While not a performance metric, it serves as a diagnostic indicator of the agent’s reasoning depth and adaptivity.
\end{description}



\section{Results}
\label{sec:results}

\subsection{Quantitative performance}
\label{subsec:results_quantitative}

Our primary experiment evaluates the \textbf{RL-Tuned Iterative Agent} against increasingly capable baselines. As shown in Table~\ref{tab:main_results}, performance steadily improves with greater interactivity. Relying on parametric knowledge (\textbf{B1}) yields 16.3\% accuracy. A single SPARQL query (\textbf{B2}) improves this to 19.7\%, though hampered by a low 47.7\% executability rate. The prompt-guided iterative agent (\textbf{B3}) demonstrates the value of a refinement loop, reaching 32.2\% accuracy.

Our \textbf{RL-Tuned Agent} marks a transformative leap, achieving a final accuracy of \textbf{49.7\%}---an absolute improvement of 17.5 percentage points over the strongest baseline. This gain is driven by a learned policy for interaction, evidenced by the executability rate soaring to 81.0\%. The improvement is statistically significant, confirmed by McNemar's test on the discordant pairs ($n_\text{RL-correct, baseline-wrong}=354$ vs. $n_\text{RL-wrong, baseline-correct}=130$), yielding $\chi^2(1) = 102.75, p \ll .001$.

\begin{table*}[t!]
    \centering
    \caption{End-to-end performance on the curated LC-QuAD 2.0 test set (N=1,279). Our RL-Tuned agent outperforms all zero-shot baselines, demonstrating the effectiveness of learning from interaction.}
    \label{tab:main_results}
    \renewcommand{\arraystretch}{1.2}
    \begin{tabular}{@{}lccc@{}}
        \toprule
        \textbf{Model / Approach} & \textbf{Exec. Rate (\%)} & \textbf{Acc. (\%)} & \textbf{Pass@5 (\%)} \\
        \midrule
        \multicolumn{4}{l}{\textit{Parametric Baseline (No KG Interaction)}} \\
        \quad B1: Direct QA (CoT) & - & 16.3 & 35.1 \\
        \midrule
        \multicolumn{4}{l}{\textit{SPARQL-based Baselines (Zero-Shot)}} \\
        \quad B2: One-Shot SPARQL & 47.7 & 19.7 & 47.2 \\
        \quad B3: Prompt-Guided Agent & 54.7 & 32.2 & 61.7 \\
        \midrule
        \multicolumn{4}{l}{\textit{Our Method (RL Fine-Tuned)}} \\
        \quad \textbf{RL-Tuned Agent} & \textbf{81.0} & \textbf{49.7} & \textbf{77.7} \\
        \bottomrule
    \end{tabular}
    \renewcommand{\arraystretch}{1.0}
\end{table*}

\subsection{Ablation: deconstructing agent performance}
\label{subsec:ablation_study}
To isolate the sources of this gain, we trained a purely \textbf{Reactive} agent (no \texttt{\thinktag} block) with the same RL process. Table~\ref{tab:ablation_results_trained} shows this agent still achieved 48.1\% accuracy, confirming that outcome-driven RL is the primary engine of performance, capable of learning effective strategies from interaction alone. However, our main \textbf{Deliberative} agent performed best (49.7\%), suggesting the \texttt{\thinktag} block acts as a powerful cognitive scaffold. By prompting the model to externalize its plan, the structure regularizes the learning process, leading to a more precise final policy.

\begin{table}[h!]
    \centering
    \caption{Performance of Deliberative vs. Reactive agents. Both were trained for one epoch with GRPO.}
    \label{tab:ablation_results_trained}
    \renewcommand{\arraystretch}{1.2}
    \begin{tabular}{@{}lcc@{}}
        \toprule
        \textbf{Model / Approach} & \textbf{Exec. Rate (\%)} & \textbf{Accuracy (\%)} \\
        \midrule
        \quad Agent-RL (Reactive) & 82.3 & 48.1 \\
        \quad \textbf{Agent-RL (Deliberative)} & \textbf{81.0} & \textbf{49.7} \\
        \bottomrule
    \end{tabular}
    \renewcommand{\arraystretch}{1.0}
\end{table}

\subsection{Analysis of learning dynamics}
\label{subsec:results_dynamics}
\begin{figure*}[t!]
    \centering
    \begin{subfigure}[b]{0.48\textwidth}
        \centering
        \includegraphics[width=\textwidth]{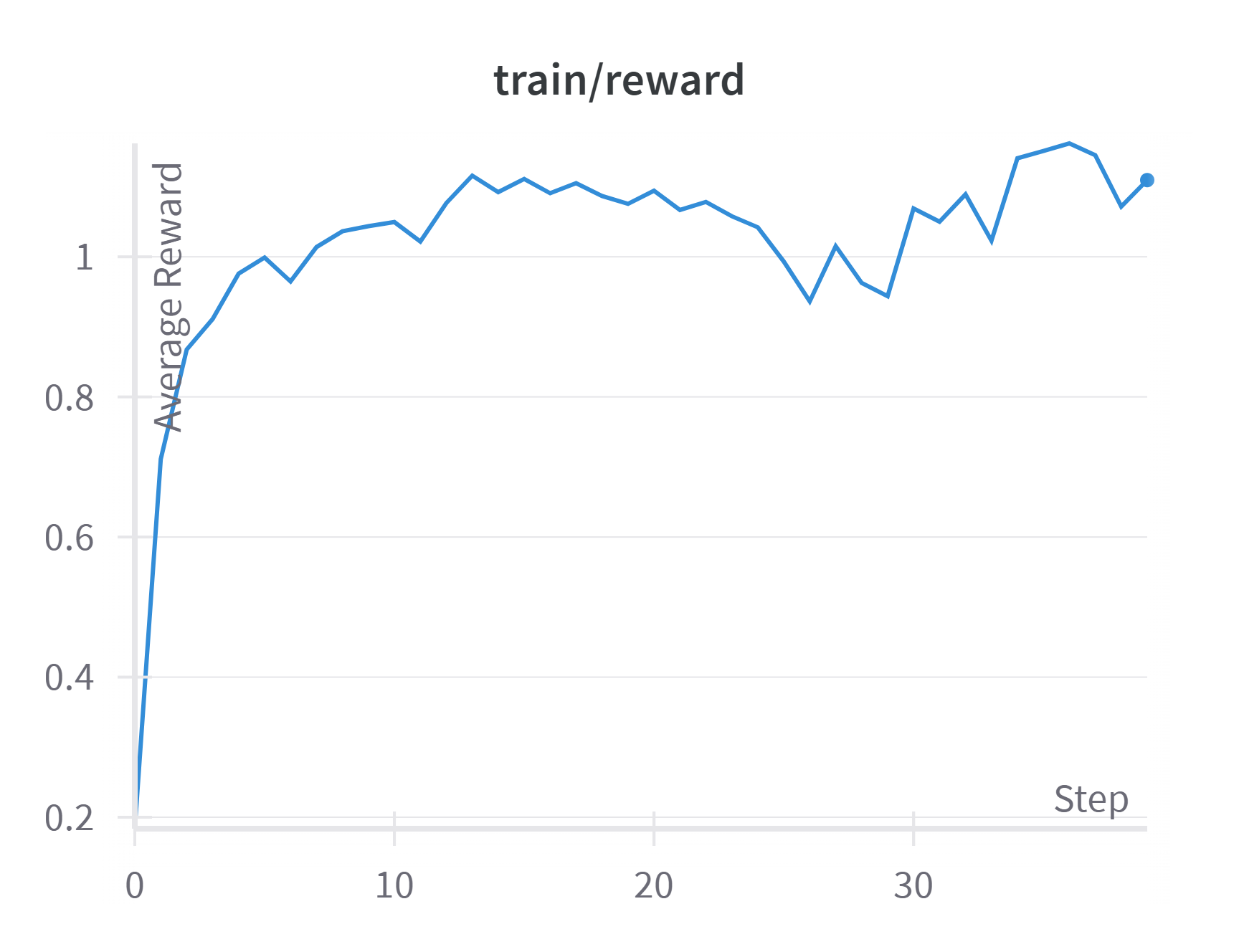}
        \caption{Smoothed Average Reward}
        \label{fig:reward_curve}
    \end{subfigure}
    \hfill
    \begin{subfigure}[b]{0.48\textwidth}
        \centering
        \includegraphics[width=\textwidth]{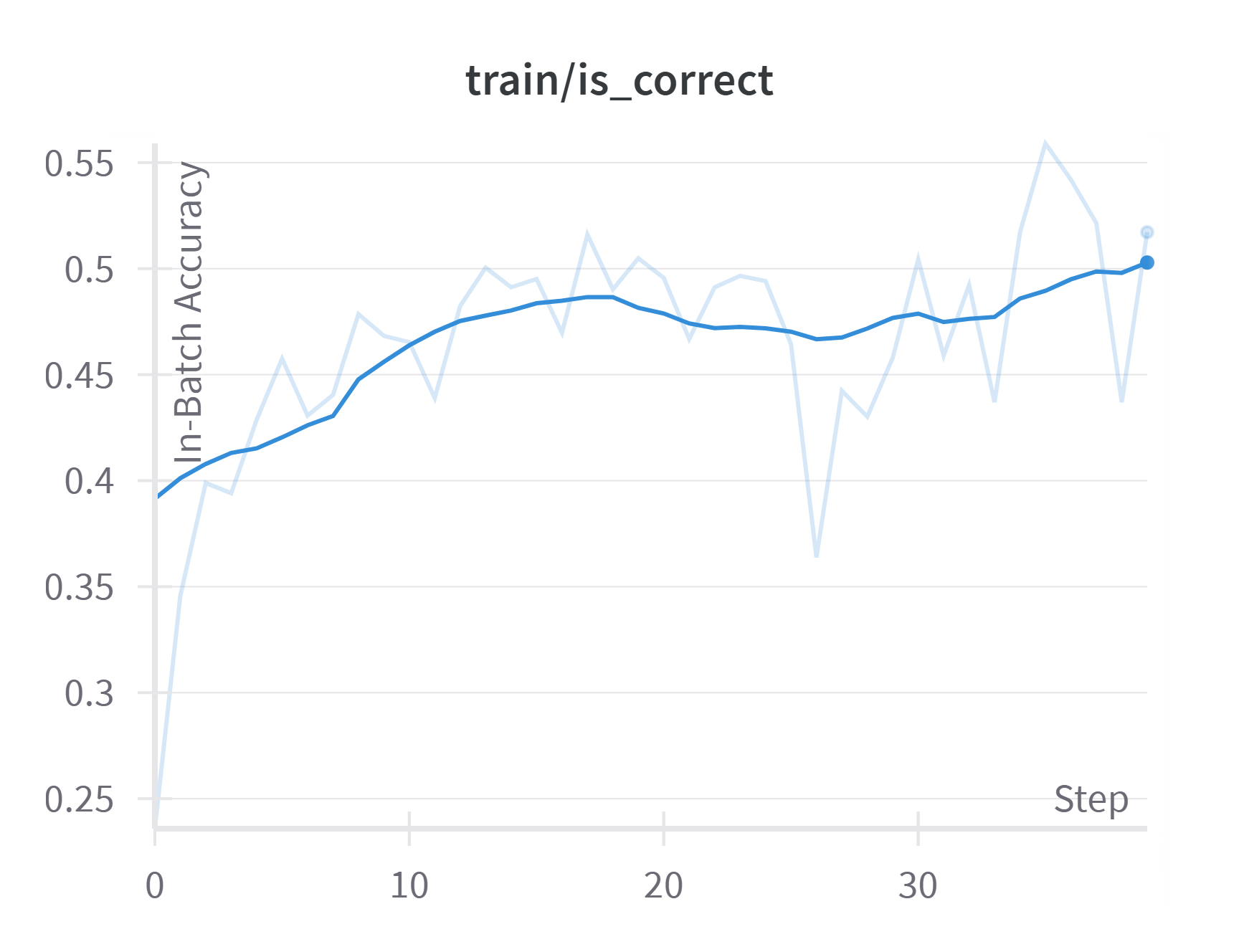}
        \caption{In-Batch Accuracy (\%)}
        \label{fig:accuracy_curve}
    \end{subfigure}
    \vspace{1em}
    \begin{subfigure}[b]{0.48\textwidth}
        \centering
        \includegraphics[width=\textwidth]{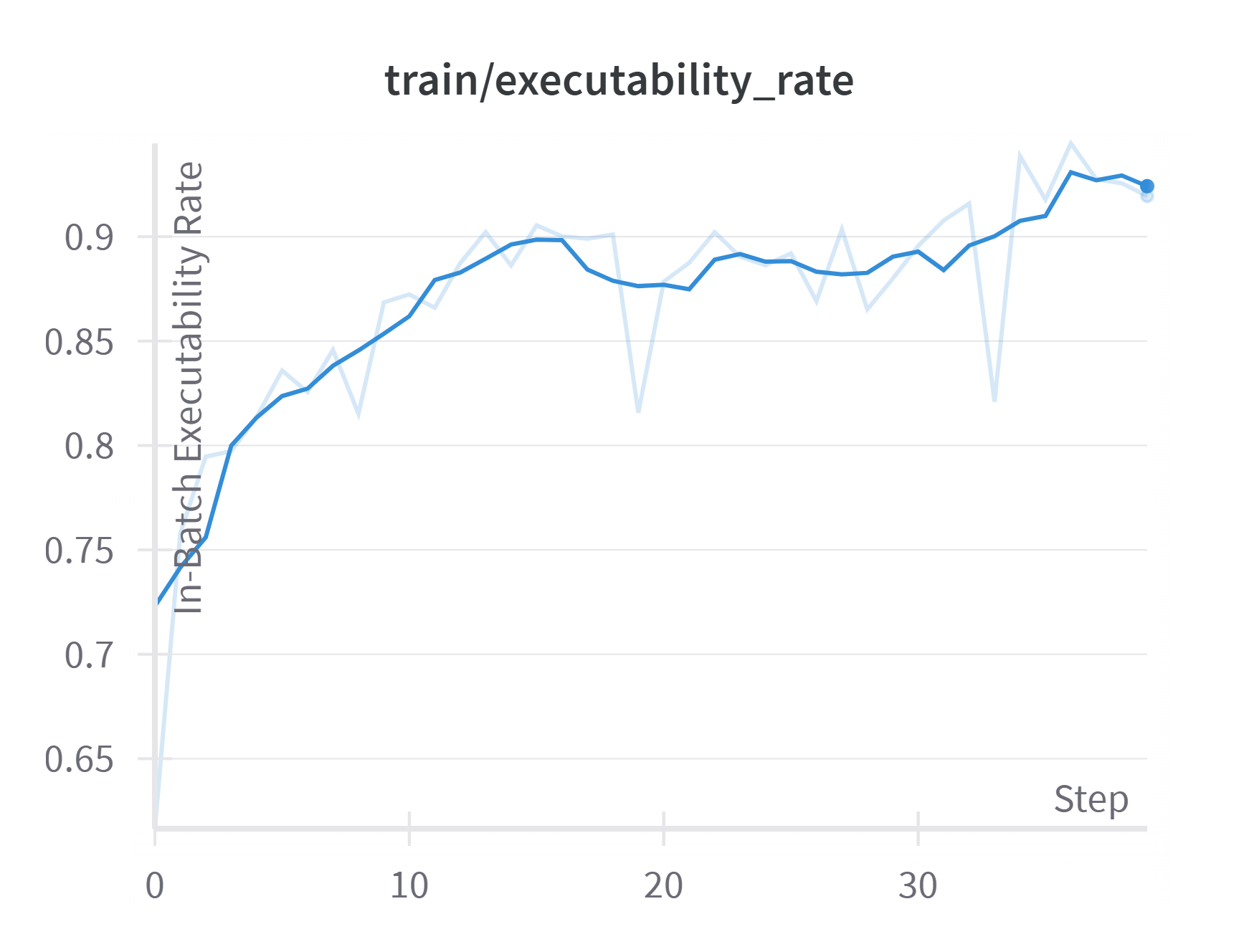}
        \caption{In-Batch Executability Rate (\%)}
        \label{fig:executability_curve}
    \end{subfigure}
    \hfill
    \begin{subfigure}[b]{0.48\textwidth}
        \centering
        \includegraphics[width=\textwidth]{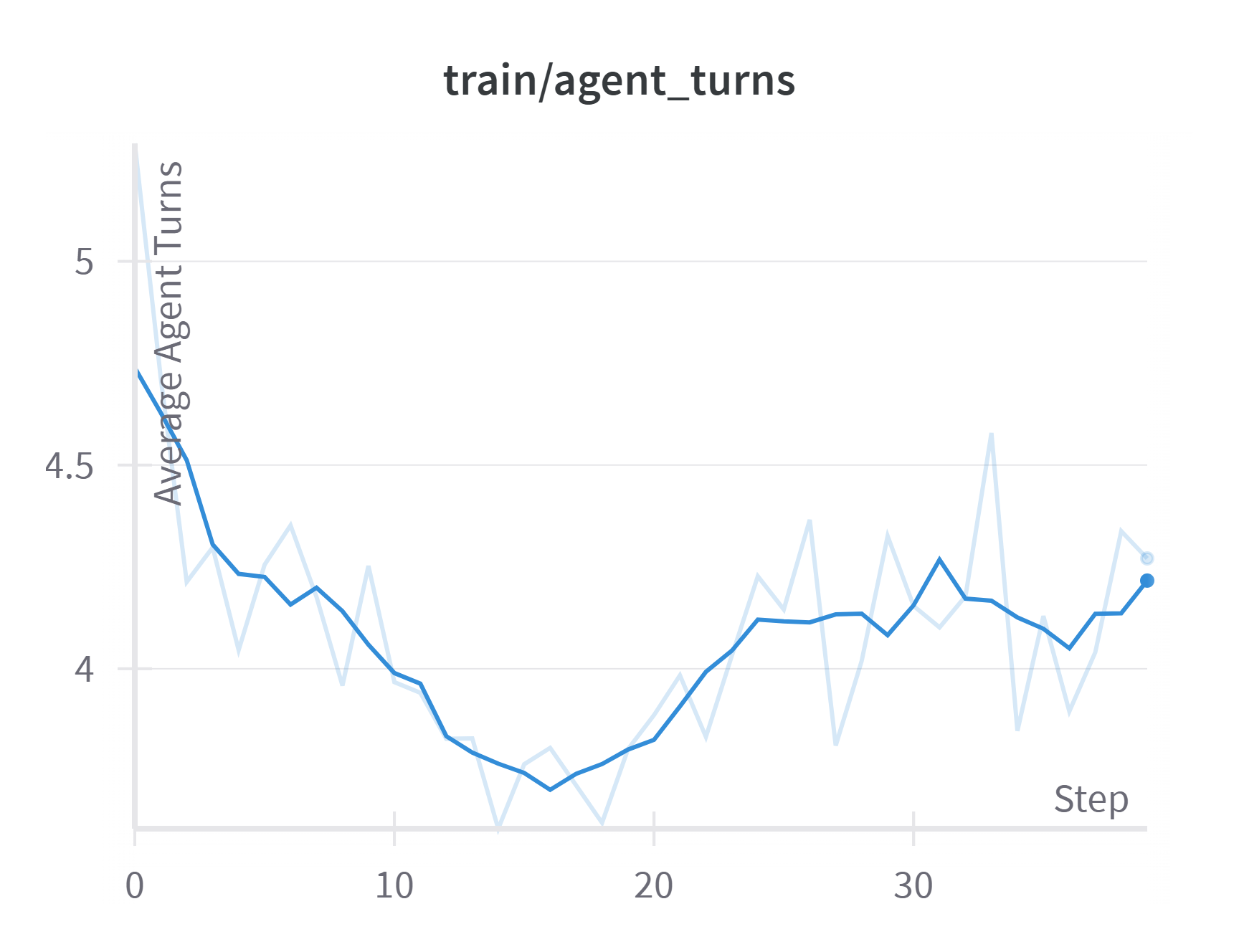}
        \caption{Average Agent Turns}
        \label{fig:agent_turns_curve}
    \end{subfigure}
    \caption{Training dynamics of the RL-Tuned Agent over one epoch (40 training steps). The plots illustrate a layered learning process: the agent is optimized for reward (a), which drives improvements first in query syntax (c) and then in semantic accuracy (b). Simultaneously, the agent learns task efficiency, reducing the average number of turns required to find an answer (d).}
    \label{fig:training_dynamics_grid}
\end{figure*}

Figure~\ref{fig:training_dynamics_grid} shows the agent learns its policy in layers. Optimizing for reward (a) first drives mastery of SPARQL syntax, as the executability rate (c) rapidly saturates. This foundational skill then enables the agent to improve its semantic reasoning, reflected in the steady rise of in-batch accuracy (b). Crucially, this improved performance is matched by greater efficiency; the average number of turns required per question trends downward (d). The agent learns to be more direct and effective, not just to succeed through brute-force trial and error.

\subsection{Qualitative and error analysis}
\label{subsec:results_qualitative}

\subsubsection{Error analysis: a shift to higher-quality failures.}
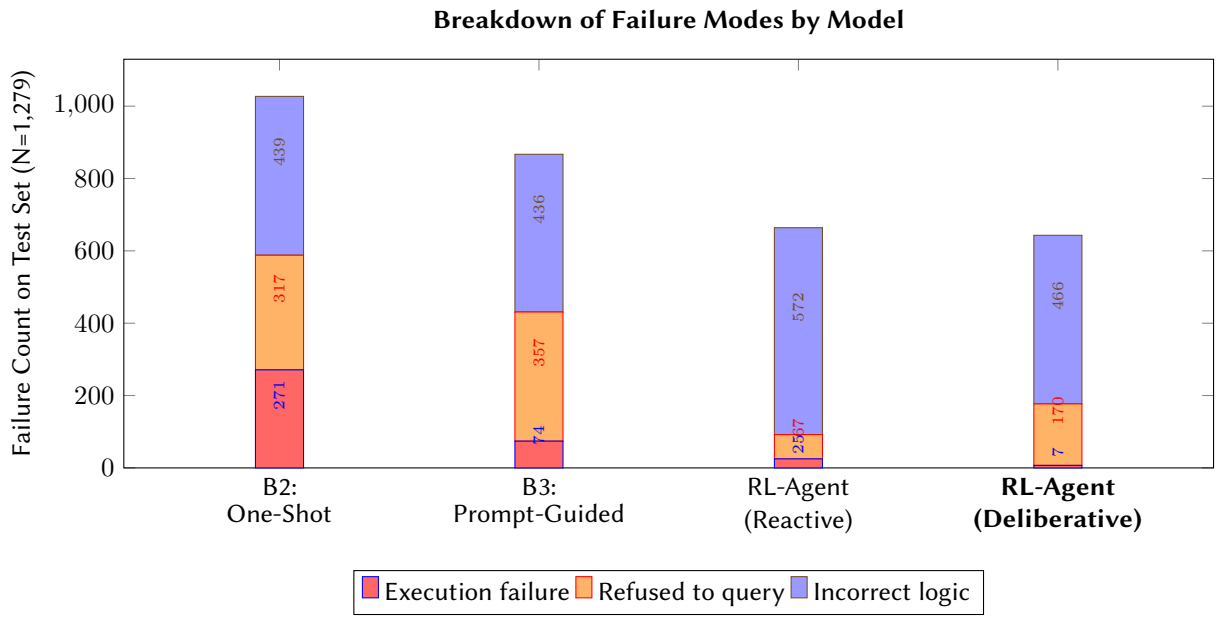
\begin{figure}[h!]
  \centering
  \begin{tikzpicture}
    \begin{axis}[
      width=\columnwidth,
      height=7cm,
      title={\textbf{Breakdown of Failure Modes by Model}},
      ybar stacked,
      bar width=18pt,
      enlarge x limits=0.2,
      ylabel={Failure Count on Test Set (N=1,279)},
      symbolic x coords={%
        B2: One-Shot,%
        B3: Prompt-Guided,%
        RL-Reactive,%
        RL-Tuned%
      },
      xtick=data,
      xticklabels={
        \shortstack{B2:\\One-Shot},
        \shortstack{B3:\\Prompt-Guided},
        \shortstack{RL-Agent\\(Reactive)},
        \shortstack{\textbf{RL-Agent}\\\textbf{(Deliberative)}}
      },
      xticklabel style={
        font=\small,
        align=center
      },
      nodes near coords,
      every node near coord/.style={
        font=\tiny,
        rotate=90,
        anchor=west,
        /pgf/number format/fixed,
      },
      legend style={
        at={(0.5,-0.25)},
        anchor=north,
        legend columns=-1,
        font=\small
      },
      clip=false,
      ymin=0,
    ]
      \addplot+[ybar, fill=red!60] coordinates {
        (B2: One-Shot, 271)
        (B3: Prompt-Guided, 74)
        (RL-Reactive, 25)
        (RL-Tuned, 7)
      };
      \addplot+[ybar, fill=orange!60] coordinates {
        (B2: One-Shot, 317)
        (B3: Prompt-Guided, 357)
        (RL-Reactive, 67)
        (RL-Tuned, 170)
      };
      \addplot+[ybar, fill=blue!40] coordinates {
        (B2: One-Shot, 439)
        (B3: Prompt-Guided, 436)
        (RL-Reactive, 572)
        (RL-Tuned, 466)
      };
      \legend{Execution failure, Refused to query, Incorrect logic}
    \end{axis}
  \end{tikzpicture}
  \caption{Absolute counts of failure modes on the test set. Reinforcement learning dramatically reduces fundamental errors like execution failures and refusal to query. This shifts the primary challenge for the trained agents from generating valid syntax to formulating correct logical plans.}
  \label{fig:error_analysis}
\end{figure}

Reinforcement learning induces a crucial shift from syntactic incompetence to semantic reasoning errors (Figure~\ref{fig:error_analysis}). The zero-shot baselines were plagued by fundamental failures: 57\% of the One-Shot agent's failures were due to execution errors or refusing to generate a query at all. In stark contrast, our RL-tuned agent nearly eliminated these issues, with such errors accounting for a negligible fraction of failures. Its primary failure mode became \textit{Incorrect Logic} (72.5\% of its own failures). The baselines fail because they cannot "speak SPARQL" correctly; our agent has mastered the tool's language and now fails on the much harder problem of reasoning correctly with it.

\subsubsection{Case study: learned resilience and strategic decomposition.}
\label{subsubsec:case_study_resilience}
To illustrate the learned policy, we analyzed behavior on the complex question: \enquote{Name the Han dynasty capital city with a twin town called Plovdiv.}. A direct query fails. Our RL-Tuned agent correctly diagnosed this, pivoted to an exploratory query to find all cities twinned with Plovdiv, and then used a second verification query on the candidates to find the correct answer. This dynamic decomposition, a direct result of RL training, contrasts sharply with the baseline, which became trapped in syntax and logic errors. 
\section{Discussion and Conclusion}
\label{sec:discussion-conclusion}
This work demonstrates that outcome-driven reinforcement learning (RL) enables compact language models to learn robust, multi-hop reasoning strategies over knowledge graphs. Our agent, trained using GRPO, significantly outperforms static and zero-shot baselines—closing the gap between symbolic structure and neural flexibility. Beyond raw accuracy, we observe emergent behaviors such as adaptive “compute scaling” and strategic query decomposition, showing that the agent learns to allocate effort based on task complexity.

These findings suggest that even small LLMs, when trained with structured feedback, can learn to navigate symbolic environments through interaction. While preliminary, this work highlights a promising direction for combining language models and formal reasoning in a more adaptive and interpretable way.

Our evaluation was limited to a curated subset of LC-QuAD 2.0 with gold entity links and single-answer queries. We did not address open-domain entity linking, incomplete or noisy KGs, or more complex answer types such as lists or aggregations. Additionally, while our approach is lightweight in model size, training with reinforcement learning remains computationally demanding. We conducted experiments on a single H100 GPU, which limits our ability to assess scalability. Questions around energy efficiency, training cost, and feasibility for broader deployment remain open and deserve closer attention in future work.

Future work includes combining supervised fine-tuning with RL to reduce sample complexity; extending to end-to-end KGQA by integrating a learned entity linker; adapting the framework to other structured domains such as NL2SQL; and studying how the agent’s policy complexity scales with model size and query difficulty.

\section*{Declaration on Generative AI}
During the preparation of this work, the author(s) used GPT-4o in order to: 
(i) check grammar and spelling, 
(ii) assist with LaTeX formatting, and 
(iii) provide sparring and critical feedback on sections. 
After using this tool, the author(s) reviewed and edited the content as needed and take full responsibility for the publication’s content.

\bibliography{references}
\end{document}